%% file: emnlp2023.tex
\pdfoutput=1

\documentclass[11pt]{article}

\usepackage{EMNLP2023}

\usepackage{times}
\usepackage{latexsym}

\usepackage[T1]{fontenc}

\usepackage[utf8]{inputenc}

\usepackage{microtype}

\usepackage{inconsolata}

\usepackage{graphicx}
\usepackage{multirow}
\usepackage{multicol}
\usepackage{subcaption}
\usepackage{enumitem,kantlipsum}
\usepackage[normalem]{ulem}
\usepackage{tabularx}
\usepackage{color, colortbl}
\usepackage{todonotes} 
\usepackage{rotating}
\usepackage{pbox}
\usepackage{booktabs} 
\usepackage{amssymb}
\usepackage{pifont}
\newcommand{\cmark}{\ding{51}}

\usepackage{tabu}
\usepackage{fontawesome}

\newcommand{\areval}{ArAIEval}

\usepackage{arabtex} 
\usepackage{utf8}
\setcode{utf8}
\title{ArAIEval Shared Task: Persuasion Techniques and\\ Disinformation Detection in Arabic Text}

\author{Maram Hasanain$^1$, Firoj Alam$^1$, Hamdy Mubarak$^1$, Samir Abdaljalil$^1$, \\
\textbf{Wajdi Zaghouani$^2$, Preslav Nakov$^3$,} \\
\textbf{Giovanni Da San Martino$^4$, Abed Alhakim Freihat$^5$}\\
$^1$Qatar Computing Research Institute, HBKU, Qatar \\
$^2$Hamad Bin Khalifa University, Qatar, \\
$^3$Mohamed bin Zayed University of Artificial Intelligence, UAE, \\  
$^4$University of Padova, Italy, 
$^5$University of Trento, Italy\\  
\texttt{\{mhasanain, fialam, hmubarak, 
 wzaghouani\}@hbku.edu.qa},  \\
\texttt{preslav.nakov@mbzuai.ac.ae}, 
\texttt{dasan@math.unipd.it},
\texttt{abdel.fraihat@gmail.com}  
\\}

\begin{document}
\maketitle
\begin{abstract}
 We present an overview of the \areval{} shared task, organized as part of the first ArabicNLP 2023 conference co-located with EMNLP 2023. \areval{} offers two tasks over Arabic text: ({\em i})~persuasion technique detection, focusing on identifying persuasion techniques in tweets and news articles, and ({\em ii})~disinformation detection in binary and multiclass setups over tweets. A total of 20 teams participated in the final evaluation phase, with 14 and 16 teams participating in Tasks 1 and 2, respectively. Across both tasks, we observed that fine-tuning transformer models such as AraBERT was at the core of the majority of the participating systems. We provide a description of the task setup, including a description of the dataset construction and the evaluation setup. We further give a brief overview of the participating systems. All datasets and evaluation scripts from the shared task are released to the research community.\footnote{\url{https://araieval.gitlab.io/}} We hope this will enable further research on these important tasks in Arabic. 
\end{list}
\end{abstract}

\input{sections/introduction}
\input{sections/related_work}
\input{sections/task1}
\input{sections/task2}

\input{sections/participant_systems}
\input{sections/conclusion}

\section*{Acknowledgments}


M. Hasanain's contribution was funded by NPRP grant 14C-0916-210015, and 
W. Zaghouani contribution was partially funded by NPRP grant 13S-0206-200281 
Both projects are funded by the Qatar National Research Fund (a member of Qatar Foundation). 


Part of this work was also funded by Qatar Foundation's IDKT Fund TDF 03-1209-210013: \emph{Tanbih: Get to Know What You Are Reading}.

We thank Fatema Akter and Hussein Mohsin Al-Dobashi for helping with the persuasion techniques annotations.


The findings herein are solely the responsibility of the authors.

\bibliography{bib/propaganda,bib/disinfo,bib/references,bib/participants}
\bibliographystyle{acl_natbib}




\end{document}

%% file: sections/introduction.tex
\section{Introduction}
\label{sec:introduction}

Social media has become one of the predominant communication channels for freely sharing content online. 
With this freedom, misuse has emerged, turning social media platforms into potential grounds for sharing inappropriate posts, misinformation, and disinformation~\citep{Zhou2016,alam-etal-2022-survey,ijcai2022p781}. Malicious users can disseminate disinformative content, such as hate-speech, rumors, and spam, to gain social and political agendas or to harm individuals, entities and organizations. Such content can inflame tension between different groups and ignite violence among their members, making early detection and prevention essential. 

Previous successful attempts to address such kinds of problems at a large scale over Arabic content include offensive and hate speech detection shared tasks~\cite{zampieri2020semeval,mubarak2020overview}. 

Social media content designed to promote hidden agendas is not limited to disinformation. In the past years, propaganda has been widely used as well, to influence and/or mislead the audience, which became a major concern for different stakeholders, social media platforms and government agencies. News reporting in the mainstream media also exhibits a similar phenomenon, where a variety of persuasion techniques~\cite{Miller} are used to promote a particular editorial agenda. 
To address this problem, the research area of ``computational propaganda'' has emerged aimed at automatically identify such techniques in textual, visual and multimodal (e.g., memes) content. \citet{EMNLP19DaSanMartino} curated a set of persuasion techniques, such as \textit{Loaded Language}, 
\textit{Appeal to Fear}, 
\textit{Straw Man} 
 and \textit{Red Herring}. 
 The focus of the work was mainly on textual content (i.e., newspaper articles). Following this prior work, in 2021, \citet{SemEval2021-6-Dimitrov} organized a shared task on propaganda techniques in memes. These efforts mainly focused on English. 
 To enrich the Arabic AI research, we have organized a shared task on detection of fine-grained propaganda techniques for Arabic, which attracted many participants~\cite{propaganda-detection:WANLP2022-overview}.

Following the success of our previous shared tasks~\cite{propaganda-detection:WANLP2022-overview,zampieri2020semeval,mubarak2020overview}, and given the great interest from the community in further pushing research in this domain, this year we organize the \textbf{Ar}abic \textbf{AI} \textbf{Eval}uation (\textbf{\areval}) shared task covering the following two tasks: {\em (i)} persuasion technique detection over tweets and news articles, and {\em (ii)} disinformation detection over tweets. 


This edition of the shared task has attracted wide participation. The task was run in two phases: {\em(i)} the development phase with 38 registrations, and 14 teams submitting their systems; and {\em(ii)} the evaluation phase with 25 registrations, and 20 teams submitting their systems.
In the remainder of this paper, we define each of the two tasks, describe the Arabic evaluation datasets that were manually constructed, and provide overview of participating systems and their official scores.


%% file: sections/related_work.tex
\section{Related Work}
\label{sec:related_work}

\subsection{Persuasion Techniques Detection}
The history of studying propaganda can be traced back to the 17th century, where the focus was to understand whether manipulation techniques were used during public events at theaters, festivals, and games~\cite{Margolin1979TheVR,pamphlet_casey}. Since then, the study of propaganda has spanned across various disciplines including history, journalism, political science, sociology, and psychology~\cite{jowett2018propaganda}. Different disciplines explored propaganda for varied purposes; for instance, in political science, it is studied to analyze the ideologies of practitioners and to understand the impact of information dissemination on public opinion.  

Over the last few decades, the current information ecosystem has undergone significant changes due to the emergence of social media platforms, which have become breeding grounds for the creation and dissemination of misinformation and propaganda. Consequently, there has been research aimed at understanding and automatically detecting such content by defining the rhetorical and psychological techniques employed on online platforms. 

Most computational approaches for automatic detection involve identifying whether textual content contains propaganda~\cite{BARRONCEDENO20191849}, identifying propagandistic techniques~\cite{Habernal.et.al.2017.EMNLP,Habernal2018b}, and detecting propagandistic text spans in news articles~\cite{EMNLP19DaSanMartino,da2020semeval}. The majority of these studies have primarily focused on English. To address this issue in multilingual settings, a shared task was recently organized, focusing on nine languages~\cite{piskorski-etal-2023-semeval}. The outcomes of such initiatives highlight the importance of multilingual models. For instance, \citet{hasanain-etal-2023-qcri} show that multilingual models significantly outperform monolingual models, even for languages unseen during training.

Other relevant shared tasks include those focusing on multimodality. \citet{SemEval2021-6-Dimitrov} organized SemEval-2021 Task 6 on the propaganda detection in memes, which comprises a multimodal setup involving both text and images.

Along such initiatives, we have primarily focused on Arabic content. The propaganda shared task, co-located with WANLP 2022, was mainly focused on tweets in both binary and multilabel settings \cite{propaganda-detection:WANLP2022-overview}. This year, we have expanded it on a larger scale with a larger dataset, focusing on news articles and tweets.

\subsection{Disinformation Detection}
\emph{\textbf{Disinformation}} is relatively a new term and it is defined as ``\emph{fabricated or deliberately manipulated text/speech/visual context, and also intentionally created conspiracy theories or rumors}'' \cite{ireton2018journalism}. There have been several studies on the automatic detection of bad content on social media, including hate speech~\cite{fortuna2018survey}, harmful content~\cite{alam-etal-2021-fighting-covid,alam-etal-2022-survey}, rumors~\cite{meel2020fake}, and offensive language~\cite{husain2021survey}. 

In the context of Arabic social media, numerous researchers have employed different approaches to disinformation detection. For instance, \citet{Boulouard22} investigated disinformation detection, particularly hate-speech and offensive content detection, on Arabic social media. 

For this shared task on disinformation detection, our work is inspired by \citet{mubarak2023detecting}, which primarily focused on detecting disinformative tweets that are most likely to be deleted. 

%% file: sections/task1.tex
\section{Task 1: Propaganda Detection}
\label{sec:task1}
The goal of this task is to identify the persuasion techniques present in a piece of text. It targets multi-genre content, including tweets and paragraphs from news articles, as persuasion techniques are commonly used within these domains. The task is organized into two subtasks.  
\subsection{Subtasks}
\label{ssec:task_definition_t1}
\paragraph{Subtask 1A:} Given a text snippet, identify whether it contains content with any persuasion technique. This is a \textit{binary classification} task.
\paragraph{Subtask 1B:} Given a text snippet, identify the propaganda techniques used in it. This is a \textit{multilabel classification} task.

\begin{table}[t]
\centering
\resizebox{0.85\columnwidth}{!}{%
 \begin{tabular}{l|lll}
\toprule
 & {\textbf{Train}} & {\textbf{Dev}} & {\textbf{Test}} \\ \cline{2-4} 
true & 1918 (79\%) & 202 (78\%)    & 331 (66\%)  \\
false & 509 (21\%)  & 57 (22\%) & 172 (34\%)  \\ \hline
\textbf{Total} & \textbf{2427}   & \textbf{259}  & \textbf{503}  \\           \toprule
\end{tabular}%
}
\caption{Distribution of Subtask \textbf{1A} dataset. In parentheses, we show the percentage of a label in a split. }
\label{tab:dataset_task1a}
\end{table}

\subsection{Dataset}
To construct the annotated dataset for this task, we collected different datasets consisting of tweets and news articles, as discussed below.

\noindent\textbf{Tweets:}
We start from the same tweets dataset collected from Twitter accounts of Arabic news sources, as described in the previous edition of the shared task~\cite{propaganda-detection:WANLP2022-overview}. We randomly sampled a subset of 156 tweets for annotation to construct the \textit{testing subset} of this task. The number of tweets selected for annotation was decided based on time and cost required for annotation.

\noindent\textbf{News paragraphs:}
We select news articles from an existing dataset, AraFacts~\cite{ali2021arafacts}, that contains claims verified by Arabic fact-checking websites, and each claim is associated with web pages propagating or negating the claim. We keep the pages that are from news domains in the set (e.g., www.alquds.co.uk). 
We automatically parsed these news articles 
and split them 
into paragraphs based on blank lines. 

\begin{table}[t]
\centering
\resizebox{\columnwidth}{!}{%
\begin{tabular}{l|lll}
\toprule
\textbf{Persuasion Technique}  & \textbf{\begin{tabular}[c]{@{}l@{}}Train \\ (2427)\end{tabular}} & \textbf{\begin{tabular}[c]{@{}l@{}}Dev \\ (259)\end{tabular}} & \textbf{\begin{tabular}[c]{@{}l@{}}Test \\ (503)\end{tabular}} \\ \hline
Loaded Language & 1574    & 176  & 253   \\
Name Calling or Labelling  & 692 & 77   & 133   \\
Questioning the Reputation & 383 & 43   & 89 \\
Exaggeration or Minimisation   & 292 & 33   & 40 \\
\begin{tabular}[c]{@{}l@{}}Obfuscation, Intentional \\ Vagueness, Confusion\end{tabular} & 240 & 28   & 25 \\
Casting Doubt   & 143 & 16   & 21 \\
Causal Oversimplification  & 128 & 15   & 12 \\
Appeal to Fear, Prejudice  & 108 & 12   & 15 \\
Slogans & 70  & 8    & 25 \\
Flag Waving & 63  & 7    & 25 \\
Appeal to Hypocrisy & 56  & 7    & 17 \\
Appeal to Values    & 37  & 4    & 29 \\
Appeal to Authority & 48  & 5    & 14 \\
False Dilemma or No Choice & 32  & 3    & 6  \\
Consequential Oversimplification   & 33  & 3    & 3  \\
Conversation Killer & 28  & 3    & 7  \\
Repetition  & 25  & 3    & 6  \\
Guilt by Association       & 13  & 1    & 1  \\
Appeal to Time  & 10  & 2    & 2  \\
Whataboutism & 9   & 1    & 2  \\
Red Herring & 8   & 1    & 3  \\
Strawman    & 6   & 1    & 2  \\
Appeal to Popularity       & 2   & 1    & 1  \\
\textit{No Technique}      & \textit{509} & \textit{57}  & \textit{172}  \\ \hline
\textbf{Total}  & \textbf{4509}   & \textbf{507} & \textbf{903}  \\
\toprule
\end{tabular}%
}
\caption{Distribution of the techniques for the Subtask \textbf{1B} dataset: sorted by total frequency over all splits. In parentheses, we show the total number of documents in a split. }
\label{tab:dataset_task1b}
\end{table}

\begin{table*}[t]
  \centering
\resizebox{0.85\textwidth}{!}{%
\begin{tabular}{ll|cc|rrrrrlrr|rrrr}
\toprule
    \multicolumn{2}{c|}{\textbf{Team}} & \multicolumn{2}{c|}{\textbf{Subtask}} & \multicolumn{8}{c|}{\textbf{Model}} & \multicolumn{4}{c}{\textbf{Misc.}} \\
    & & \multicolumn{1}{l}{\begin{sideways}1A\end{sideways}} & \multicolumn{1}{l|}{\begin{sideways}1B\end{sideways}} & \multicolumn{1}{l}{\begin{sideways}AraBERT\end{sideways}} & \multicolumn{1}{l}{\begin{sideways}MARBERT\end{sideways}} & \multicolumn{1}{l}{\begin{sideways}ArabicBERT\end{sideways}} & \multicolumn{1}{l}{\begin{sideways}BERT\end{sideways}} & \multicolumn{1}{l}{\begin{sideways}RoBERTa\end{sideways}} & \begin{sideways}XLM-RoBERTa\end{sideways} & \multicolumn{1}{l}{\begin{sideways}AraELECTRA\end{sideways}} & \multicolumn{1}{l|}{\begin{sideways}GPT\end{sideways}} & \multicolumn{1}{l}{\begin{sideways}Data augm.\end{sideways}} & \multicolumn{1}{l}{\begin{sideways}Preprocessing\end{sideways}} & \multicolumn{1}{l}{\begin{sideways}Ensamble\end{sideways}} & \multicolumn{1}{l}{\begin{sideways}Loss Funct.\end{sideways}} \\
    \midrule
    \rowcolor[rgb]{ .851,  .851,  .851} HTE   & \cite{araieval-arabicnlp:2023:task1A:HTE} & 1     & 5     & \multicolumn{1}{l}{\cmark} & \multicolumn{1}{l}{\cmark} & & & & & & & & & & \multicolumn{1}{l}{\cmark} \\
    KnowTellConvince & \cite{araieval-arabicnlp:2023:taskboth:KnowTellConvince} & 2     & & & & \multicolumn{1}{l}{\cmark} & & & & & & \multicolumn{1}{l}{\cmark} & & \multicolumn{1}{l}{\cmark} & \multicolumn{1}{l}{\cmark} \\
    \rowcolor[rgb]{ .851,  .851,  .851} rematchka & \cite{araieval-arabicnlp:2023:bothtasks:rematchka} & 3     & 2     & \multicolumn{1}{l}{\cmark} & \multicolumn{1}{l}{\cmark} & & & & & & & & & & \multicolumn{1}{l}{\cmark} \\
    UL \& UM6P & \cite{araieval-arabicnlp:2023:task:UL_UM6P} & 4     & 1     & \multicolumn{1}{l}{\cmark} & \multicolumn{1}{l}{\cmark} & & & & & & & & & & \multicolumn{1}{l}{\cmark} \\
    \rowcolor[rgb]{ .851,  .851,  .851} Itri Amigos & \cite{araieval-arabicnlp:2023:task:Itri_Amigos} & 5     & 4     & \multicolumn{1}{l}{\cmark} & & & & & & & & & & &  \\
    Raphael & \cite{araieval-arabicnlp:2023:task1:raphael} & 6     & 6     & & \multicolumn{1}{l}{\cmark} & & & \multicolumn{1}{l}{\cmark} & & & \multicolumn{1}{l|}{\cmark} & & & &  \\
    \rowcolor[rgb]{ .851,  .851,  .851} Frank & \cite{araieval-arabicnlp:2023:bothtasks:frank} & 7     & & & \multicolumn{1}{l}{\cmark} & & \multicolumn{1}{l}{\cmark} & \multicolumn{1}{l}{\cmark} & & & & & \multicolumn{1}{l}{\cmark} & &  \\
    Mavericks & \cite{araieval-arabicnlp:2023:bothtasks:Mavericks} & 8     & & \multicolumn{1}{l}{\cmark} & & & & & & \multicolumn{1}{l}{\cmark} & & & \multicolumn{1}{l}{\cmark} & \multicolumn{1}{l}{\cmark} &  \\
    \rowcolor[rgb]{ .851,  .851,  .851} Nexus & \cite{araieval-arabicnlp:2023:bothtasks:nexus} & 9     & & \multicolumn{1}{l}{\cmark} & \multicolumn{1}{l}{\cmark} & & & & & & & & \multicolumn{1}{l}{\cmark} & &  \\
    AAST-NLP & \cite{araieval-arabicnlp:2023:task:AAST-NLP} & 11    & 3     & \multicolumn{1}{l}{\cmark} & \multicolumn{1}{l}{\cmark} & & & & & & & \multicolumn{1}{l}{\cmark} & \multicolumn{1}{l}{\cmark} & & \multicolumn{1}{l}{\cmark} \\
    \rowcolor[rgb]{ .851,  .851,  .851} ReDASPersuasion & \cite{araieval-arabicnlp:2023:task:redaspersuasion} & 13    & 7     & & & & & & \cmark & & & & \multicolumn{1}{l}{\cmark} & &  \\
    Legend & \cite{araieval-arabicnlp:2023:task:legend} & 14    & & & & & & & \cmark & & & & & &  \\
    \toprule
    \end{tabular}%
    }
\caption{Overview of the systems for \textbf{Task~1}. Numbers under the subtask code indicate the position of the team in the official ranking. Data augm.: Data augmentation. Loss Funct.: Experiments with a variety of loss functions.}
\label{tab:overview_task1}    
\end{table*}

\begin{table*}[t!]
\centering
\resizebox{0.85\textwidth}{!}{%
\begin{tabular}{rlcc|rlcc}
\hline
\multicolumn{1}{l}{} & \textbf{Team}  & \multicolumn{1}{l}{\textbf{Micro F1}}   & \multicolumn{1}{l|}{\textbf{Macro F1}}  & \multicolumn{1}{l}{} & \textbf{Team}  & \multicolumn{1}{l}{\textbf{Micro F1}}   & \multicolumn{1}{l}{\textbf{Macro F1}}   \\ \hline
\multicolumn{4}{c|}{\textbf{Subtask 1A}}  & \multicolumn{4}{c}{\textbf{Subtask 1B}} \\ \hline
1 & HTE & 0.7634 & 0.7321 & 1 & UL \& UM6P & 0.5666 & 0.2156    \\
2 & KnowTellConvince & 0.7575 & 0.7282 & 2 & rematchka  & 0.5658 & 0.2497    \\
3 & rematchka & 0.7555 & 0.7309 & 3 & AAST-NLP   & 0.5522 & 0.1425    \\
4 & UL \& UM6P & 0.7515 & 0.7186 & 4 & Itri Amigos    & 0.5506 & 0.1839    \\
5 & Itri Amigos    & 0.7495 & 0.7225 & 5 & HTE & 0.5412 & 0.0979    \\
6 & Raphael    & 0.7475 & 0.7221 & 6 & Raphael    & 0.5347 & 0.1772    \\
7 & Frank & 0.7455 & 0.7173 & 7 & ReDASPersuasion & 0.4523 & 0.0568    \\
8 & Mavericks  & 0.7416 & 0.7031 & 8 & \textit{Baseline (Majority)} & 0.3599 & 0.0279 \\
9 & Nexus & 0.7396 & 0.6929 & 9 & \textit{Baseline (Random)}   & 0.0868 & 0.0584 \\
10 & superMario & 0.7316 & 0.7098 & 10 & pakapro    & 0.0854 & 0.0563    \\
11 & AAST-NLP   & 0.7237 & 0.6693 & \multicolumn{1}{l}{} & & \multicolumn{1}{l}{}  & \multicolumn{1}{l}{}  \\
12 & \textit{Baseline (Majority)} & 0.6581 & 0.3969 & \multicolumn{1}{l}{} & & \multicolumn{1}{l}{}  & \multicolumn{1}{l}{}  \\
13 & ReDASPersuasion & 0.6581 & 0.3969 & \multicolumn{1}{l}{} & & \multicolumn{1}{l}{}  & \multicolumn{1}{l}{}  \\
14 & Legend & 0.6402 & 0.4647 & \multicolumn{1}{l}{} & & \multicolumn{1}{l}{}  & \multicolumn{1}{l}{}  \\
15 & pakapro    & 0.5030 & 0.4940 & \multicolumn{1}{l}{} & & \multicolumn{1}{l}{}  & \multicolumn{1}{l}{}  \\
16 & \textit{Baseline (Random)}   & 0.4771 & 0.4598 & \multicolumn{1}{l}{} & & \multicolumn{1}{l}{}  & \multicolumn{1}{l}{}  \\ \hline
\end{tabular}%
}
\caption{Official results for \textbf{Task 1}. Runs ranked by the official measure: Micro F1.}
\label{tab:results_task1}
\end{table*}

\paragraph{Data annotation:}
For both tweets and paragraphs, we follow the same annotation process to identify the persuasion techniques in a snippet. The process includes two phases: (\emph{i})~three annotators independently annotated the same text snippet, through an annotation interface designed for the task,
and (\emph{ii})~two consolidators reviewed the annotations and produced the gold annotations. Annotators were recruited and trained for the task in-house. We annotate text by a set of 23 persuasion techniques that is adopted from existing research~\cite{piskorski-etal-2023-semeval}. We should note here that multiple techniques can be found in the same text snippet. \textit{For Subtask 1A (binary classification)}, the labels were generated by assigning a positive label (true) to every text snippet that had at least one persuasion technique, and a negative label was given otherwise. Below we give an example subset of the persuasion techniques, and briefly summarize them:
\begin{enumerate}[leftmargin=*]
\itemsep0em
\item \textbf{Loaded language:} using specific emotionally-loaded words or phrases (positive or negative) to convince the audience that an argument is valid.
\item \textbf{Appeal to Fear, Prejudice:} building support or rejection for an idea by instilling fear or repulsion towards it, or to an alternative idea.
\item \textbf{Strawman:} giving the impression that an argument is being refuted, whereas the real subject of the argument was not addressed or refuted, but instead was replaced with a different one.
\end{enumerate}



\paragraph{Data splits:}
The full set of annotated paragraphs is divided into three subsets: train, development, and test, using a stratified splitting approach to ensure that the distribution of persuasion techniques is consistent across the splits. For the tweets set, we split the full annotated tweet set from the previous edition of the lab~\cite{propaganda-detection:WANLP2022-overview} into train and development subsets, while the test set is annotated for this shared task. Finally, we construct the multi-genre subsets for the task by merging the sets of paragraphs and tweets.

\paragraph{Statistics:}
In Tables \ref{tab:dataset_task1a} and \ref{tab:dataset_task1b} we show the distribution of labels across splits for Task 1.

\subsection{Evaluation Setup}
\label{ssec:evaluation_setup}
The task was organized into two phases:
\begin{itemize}
    \item \textbf{Development phase}: we released the train and development subsets, and participants submitted runs on the development set through a competition on Codalab~\footnote{\url{https://codalab.lisn.upsaclay.fr/competitions/14563}}.
    \item \textbf{Test phase}: we released the official test subset, and the participants were given a few days to submit their final predictions through a competition on Codalab.\footnote{\url{https://codalab.lisn.upsaclay.fr/competitions/15099}} Only the latest submission from each team was considered official and was used for the final team ranking.
\end{itemize}

\noindent\textbf{Measures:} We measure the performance of the participating systems, for all subtasks, using micro-averaged F1 as the official evaluation measure of the shared task, as these are multiclass/multilabel problems, where the labels are imbalanced. We also report macro-averaged F1, as an unofficial evaluation measure.

\subsection{Overview of Participating Systems and Results}
\label{ssec:results}
A total of 14 and 8 teams submitted runs for Subtask 1A and 1B, respectively, with 8 teams making submissions for both subtasks. Table~\ref{tab:overview_task1}, overviews 12 of the participating systems for which a description paper was submitted. Table~\ref{tab:results_task1} presents the results and rankings of \textit{all} systems. 


Fine-tuning pre-trained Arabic models (specifically AraBERT~\cite{antoun2020arabert} and MARBERT~\cite{abdul-mageed-etal-2021-arbert}) was the most common system architecture. However, we observed that several systems also experimented with a variety of loss functions for model training to handle characteristics of the training dataset, like label imbalance~\cite{araieval-arabicnlp:2023:task:UL_UM6P,araieval-arabicnlp:2023:task1A:HTE,araieval-arabicnlp:2023:taskboth:KnowTellConvince,araieval-arabicnlp:2023:bothtasks:rematchka,araieval-arabicnlp:2023:task:AAST-NLP}.

When comparing the performance to the previous edition~\cite{propaganda-detection:WANLP2022-overview} for the multilabel subtask, we observe that this year's Subtask 1B is much more challenging. In the previous edition, the best system achieved a Micro F1 of 0.649, whereas this year it is 0.566, keeping in mind that the dataset is different and may not be exactly comparable.  

%% file: sections/task2.tex
\section{Task 2: Disinformation Detection}
\label{sec:task2}
This task targeted tweets and was organized into two subtasks, as discussed below.  

\subsection{Subtasks}
\label{ssec:task_def_t2}
\paragraph{Subtask 2A:} Given a tweet, identify whether it is disinformative. 
This is a \textit{binary classification} task.
\paragraph{Subtask 2B:} Given a tweet, detect the fine-grained disinformation class, if any. This is a \textit{multiclass classification} task. The fine-grained labels include \textit{hate-speech}, \textit{offensive}, \textit{rumor}, and \textit{spam}.



\begin{table}[t]
\centering

\resizebox{0.9\columnwidth}{!}{%
\begin{tabular}{l|lll}
\toprule
& \textbf{Train} & \textbf{Dev} & \textbf{Test} \\ \cline{2-4} 
Disinfo        & 2656 (19\%)    & 397 (19\%)   & 876 (23\%)    \\
Not-disinfo     & 11491 (81\%)   & 1718 (81\%)  & 2853 (77\%)   \\ \hline
\textbf{Total} & \textbf{14147}          & \textbf{2115}         & \textbf{3729}  \\
\toprule
\end{tabular}%
}
\caption{Distribution of Subtask \textbf{2A} dataset. In parentheses, we show the percentage of a label in a split. }
\label{tab:dataset_task2a}
\end{table}

\subsection{Dataset}
\label{ssec:dataset_t2}
We have constructed an annotated dataset composed of $20K$ tweets, labeled as disinformative or not-disinformative, along with fine-grained categories for the disinformative set. These tweets are related to COVID-19 and were collected in February and March 2020. 
 We followed the annotation guidelines described in~\cite{mubarak2020overview},~\cite{zampieri2020semeval},~\cite{mubarak2022arcovidvac}, and~\cite{mubarak2020spam}, for hate speech, offensive content, rumor, and spam classes, respectively. More details about data collection and annotation can be found in~\cite{mubarak2023detecting}. Tables \ref{tab:dataset_task2a} and \ref{tab:dataset_task2b} display the statistics of the dataset.

\begin{table}[t]
\centering
\resizebox{0.85\columnwidth}{!}{%
\begin{tabular}{l|lll}
\toprule
 & \textbf{Train} & \textbf{Dev} & \textbf{Test} \\ \cline{2-4} 
HS             & 1512 (57\%)    & 226 (57\%)   & 442 (50\%)    \\
Off            & 500 (19\%)     & 75 (19\%)    & 160 (18\%)    \\
Rumor          & 191 (7\%)      & 28 (7\%)     & 33 (4\%)      \\
Spam           & 453 (17\%)     & 68 (17\%)    & 241 (28\%)    \\ \hline
\textbf{Total} & \textbf{2656}  & \textbf{397} & \textbf{876}  \\ \toprule
\end{tabular}%
}
\caption{Distribution of Subtask \textbf{2B} dataset. In parentheses, we show the percentage of a label in a split. 
}
\label{tab:dataset_task2b}
\end{table}

\begin{table*}[t]
  \centering
\resizebox{0.85\textwidth}{!}{%
    \begin{tabular}{rrcccccccccccccccc}
\toprule
    \multicolumn{2}{c|}{\textbf{Team}} & \multicolumn{2}{c|}{\textbf{Subtask}} & \multicolumn{12}{c|}{\textbf{Model}}                                                                   & \multicolumn{2}{c}{\textbf{Misc.}} \\
          & \multicolumn{1}{r|}{} & \begin{sideways}2A\end{sideways} & \multicolumn{1}{c|}{\begin{sideways}2B\end{sideways}} & \begin{sideways}AraBERT\end{sideways} & \begin{sideways}MARBERT\end{sideways} & \begin{sideways}ARBERT\end{sideways} & \begin{sideways}QARiB\end{sideways} & \begin{sideways}CAMeLBERT\end{sideways} & \begin{sideways}BERT\end{sideways} & \begin{sideways}RoBERTa\end{sideways} & \begin{sideways}XLM-RoBERTa\end{sideways} & \begin{sideways}DistilBERT\end{sideways} & \begin{sideways}AraELECTRA\end{sideways} & \begin{sideways}LSTM\end{sideways} & \multicolumn{1}{c|}{\begin{sideways}SVM\end{sideways}} & \begin{sideways}Data augm.\end{sideways} & \begin{sideways}Preprocessing\end{sideways} \\
    \midrule
    \rowcolor[rgb]{ .851,  .851,  .851} \multicolumn{1}{l}{DetectiveRedasers} & \multicolumn{1}{l|}{\cite{araieval-arabicnlp:2023:task2:DetectiveRedasers}} & 1     & \multicolumn{1}{c|}{1} & \cmark & \cmark &       & \cmark & \cmark &       &       &       &       &       &       & \multicolumn{1}{c|}{} & \cmark & \cmark \\
    \multicolumn{1}{l}{AAST-NLP} & \multicolumn{1}{l|}{\cite{araieval-arabicnlp:2023:task:AAST-NLP}} & 2     & \multicolumn{1}{c|}{3} & \cmark & \cmark & \cmark &       &       &       &       &       &       &       & \cmark & \multicolumn{1}{c|}{} & \cmark & \cmark \\
    \rowcolor[rgb]{ .851,  .851,  .851} \multicolumn{1}{l}{UL \& UM6P} & \multicolumn{1}{l|}{\cite{araieval-arabicnlp:2023:task:UL_UM6P}} & 3     & \multicolumn{1}{c|}{2} & \cmark & \cmark & \cmark &       &       &       &       &       &       &       &       & \multicolumn{1}{c|}{} &       &  \\
    \multicolumn{1}{l}{rematchka} & \multicolumn{1}{l|}{\cite{araieval-arabicnlp:2023:bothtasks:rematchka}} & 4     & \multicolumn{1}{c|}{4} & \cmark & \cmark & \cmark &       &       &       &       &       &       &       &       & \multicolumn{1}{c|}{} & \cmark &  \\
    \rowcolor[rgb]{ .851,  .851,  .851} \multicolumn{1}{l}{PD-AR} & \multicolumn{1}{l|}{\cite{araieval-arabicnlp:2023:task:PD-AR}} & 5     & \multicolumn{1}{c|}{6} & \cmark &       &       &       & \cmark & \cmark & \cmark & \cmark &       &       &       & \multicolumn{1}{c|}{} &       & \cmark \\
    \multicolumn{1}{l}{Mavericks} & \multicolumn{1}{l|}{\cite{araieval-arabicnlp:2023:bothtasks:Mavericks}} & 7     & \multicolumn{1}{c|}{} & \cmark &       &       &       &       &       &       &       &       & \cmark &       & \multicolumn{1}{c|}{} &       & \cmark \\
    \rowcolor[rgb]{ .851,  .851,  .851} \multicolumn{1}{l}{Itri Amigos} & \multicolumn{1}{l|}{\cite{araieval-arabicnlp:2023:task:Itri_Amigos}} & 8     & \multicolumn{1}{c|}{7} & \cmark &       &       &       &       &       &       &       &       &       &       & \multicolumn{1}{c|}{} &       & \cmark \\
    \multicolumn{1}{l}{KnowTellConvince} & \multicolumn{1}{l|}{\cite{araieval-arabicnlp:2023:taskboth:KnowTellConvince}} & 9     & \multicolumn{1}{c|}{8} & \cmark &       &       &       &       &       &       &       &       &       &       & \multicolumn{1}{c|}{} &       &  \\
    \rowcolor[rgb]{ .851,  .851,  .851} \multicolumn{1}{l}{Nexus} & \multicolumn{1}{l|}{\cite{araieval-arabicnlp:2023:bothtasks:nexus}} & 10    & \multicolumn{1}{c|}{} & \cmark & \cmark &       & \cmark &       &       &       &       &       &       &       & \multicolumn{1}{c|}{} &       &  \\
    \multicolumn{1}{l}{PTUK-HULAT} & \multicolumn{1}{l|}{\cite{araieval-arabicnlp:2023:task:PTUK-HULAT}} & 11    & \multicolumn{1}{c|}{} &       &       &       &       &       & \cmark &       &       & \cmark &       &       & \multicolumn{1}{c|}{} &       & \cmark \\
    \rowcolor[rgb]{ .851,  .851,  .851} \multicolumn{1}{l}{Frank} & \multicolumn{1}{l|}{\cite{araieval-arabicnlp:2023:bothtasks:frank}} & 12    & \multicolumn{1}{c|}{} &       & \cmark &       &       &       & \cmark & \cmark &       &       &       &       & \multicolumn{1}{c|}{} &       &  \\
    \multicolumn{1}{l}{USTHB} & \multicolumn{1}{l|}{\cite{araieval-arabicnlp:2023:task2:usthb}} & 13    & \multicolumn{1}{c|}{9} &       &       &       &       &       &       &       &       &       &       &       & \multicolumn{1}{c|}{\cmark} &       &  \\
    \rowcolor[rgb]{ .851,  .851,  .851} \multicolumn{1}{l}{AraDetector} & \multicolumn{1}{l|}{\cite{araieval-arabicnlp:2023:task2:aradetector}} & 15    & \multicolumn{1}{c|}{} & \cmark & \cmark &       & \cmark &       &       &       &       &       &       &       & \multicolumn{1}{c|}{} &       & \cmark \\ \toprule
    \end{tabular}%
    }
    \caption{Overview of the systems for \textbf{Task~2}. The numbers under the subtask code indicate the position of the team in the official ranking. Data augm.: Data augmentation.}
\label{tab:overview_subtask2}

\end{table*}

\begin{table*}[t!]
\centering
\resizebox{0.85\textwidth}{!}{%
\begin{tabular}{rlcc|rlcc}
\hline
 & \multicolumn{1}{l}{\textbf{Team}}      & \multicolumn{1}{l}{\textbf{Micro F1}} & \multicolumn{1}{l|}{\textbf{Macro F1}} & & \multicolumn{1}{l}{\textbf{Team}}   & \multicolumn{1}{l}{\textbf{Micro F1}} & \multicolumn{1}{l}{\textbf{Macro F1}} \\ \hline
\multicolumn{4}{c|}{\textbf{Subtask 2A}}    & \multicolumn{4}{c}{\textbf{Subtask 2B}}                                                                                                            \\ \hline
1  & DetectiveRedasers   & 0.9048 & 0.8626  & 1   & DetectiveRedasers   & 0.8356 & 0.7541                                \\
2  & AAST-NLP     & 0.9043 & 0.8634  & 2   & UL \& UM6P   & 0.8333 & 0.7388                                \\
3  & UL \& UM6P   & 0.9040 & 0.8645  & 3   & AAST-NLP     & 0.8253 & 0.7283                                \\
4  & rematchka    & 0.9040 & 0.8614  & 4   & rematchka    & 0.8219 & 0.7156                                \\
5  & PD-AR & 0.9021 & 0.8595  & 5   & superMario   & 0.8208 & 0.7031                                \\
6  & superMario   & 0.9019 & 0.8625  & 6   & PD-AR & 0.8174 & 0.7209                                \\
7  & Mavericks    & 0.9010 & 0.8606  & 7   & Itri Amigos  & 0.8139 & 0.7220                                \\
8  & Itri Amigos  & 0.8984 & 0.8468  & 8   & KnowTellConvince & 0.8071 & 0.6888                                \\
9  & KnowTellConvince & 0.8938 & 0.8460  & 9   & USTHB & 0.5046 & 0.1677                                \\
10 & Nexus & 0.8935 & 0.8459  & 10  & \textit{Baseline (Majority)} & 0.5046 & 0.1677        \\
11 & PTUK-HULAT   & 0.8675 & 0.7992  & 11  & Ankit & 0.4167 & 0.1993                                \\
12 & Frank & 0.8163 & 0.6378  & 12  & \textit{Baseline (Random)}   & 0.2603 & 0.2243        \\
13 & USTHB & 0.7670 & 0.4418  & 13  & pakapro      & 0.2317 & 0.1978                                \\
14 & \textit{Baseline (Majority)} & 0.7651 & 0.4335  & \multicolumn{1}{l}{} &       & \multicolumn{1}{l}{} & \multicolumn{1}{l}{}                  \\
15 & AraDetector  & 0.7487 & 0.6498  & \multicolumn{1}{l}{} &       & \multicolumn{1}{l}{} & \multicolumn{1}{l}{}                  \\
16 & \textit{Baseline (Random)}   & 0.5154 & 0.4764  & \multicolumn{1}{l}{} &       & \multicolumn{1}{l}{} & \multicolumn{1}{l}{}                  \\
17 & pakapro      & 0.4996 & 0.4596  & \multicolumn{1}{l}{} &       & \multicolumn{1}{l}{} & \multicolumn{1}{l}{}\\ \hline
\end{tabular}%
}
\caption{Official results for \textbf{Task 2}. Runs ranked by the official measure: Micro F1.}
\label{tab:results_task2}

\end{table*}

\subsection{Evaluation Setup and Measures}
\label{ssec:evaluation_setup_t2}
Similar to Task 1, we also conducted this task in two phases as discussed in Section~\ref{ssec:evaluation_setup}. Systems were valuated using Micro F1 as the official measure, while also reporting Macro F1.


\subsection{Overview of Participating Systems and Results}
\label{ssec:results_t2}
Table~\ref{tab:overview_subtask2} and \ref{tab:results_task2} overviews the submitted systems, and the official results and ranking, respectively. A total of 15 and 11 teams participated in Subtask 2A and 2B, respectively, out of which, 10 made submissions for both subtasks. Out of 17 teams, 13 outperformed the majority baseline for Subtask 2A, whereas out of 11 teams, 9 outperformed the majority baseline for Subtask 2B. These subtasks were dominated by transformer models as observed in Table~\ref{tab:overview_subtask2}. The most commonly used model was AraBERT~\cite{antoun2020arabert}, followed by MARBERT~\cite{abdul-mageed-etal-2021-arbert}, ARBERT\cite{abdul-mageed-etal-2021-arbert}, and QARiB~\cite{abdelali2021pretraining}. Half of the participants employed preprocessing techniques, and the top-performing teams utilized data augmentation.

%% file: sections/participant_systems.tex
\section{Participating Systems}
\label{sec:participating_systems}


\paragraph{AAST-NLP~\cite{araieval-arabicnlp:2023:task:AAST-NLP}}
The team experimented with several transformer-based models, including MARBERT~\cite{abdul-mageed-etal-2021-arbert}, ARBERT~\cite{abdul-mageed-etal-2021-arbert}, and AraBERT~\cite{antoun2020arabert}. AraBERT outperformed the others across all subtasks. Preprocessing was applied using the AraBERT preprocessor. Tweet tags, emojis, and Arabic stopwords were removed. For the final submission, binary cross entropy was selected for multilabel classification (Subtask 1B), while Dice loss was chosen for the remaining three subtasks. Although the team tried data augmentation with contextual word embeddings and a hybrid approach combining AraBERT with a CNN-BILSTM, these did not improve accuracy.


\paragraph{AraDetector~\cite{araieval-arabicnlp:2023:task2:aradetector}} The team tackled Subtask 2A using an ensemble of three classifiers: MARBERT model fine-tuned on the training data, and GPT-4~\cite{openai2023gpt4} in zero-shot and few-shot settings. A majority voting approach was then used to merge the binary predictions of the three classifiers. The results on the development set showed that GPT-4 in zero-shot setting outperforms the ensemble model by the Micro F1 measure. 

\paragraph{DetectiveRedasers~\cite{araieval-arabicnlp:2023:task2:DetectiveRedasers} }
The team participated in subtasks 2A and 2B following a two-fold methodology. First, they conducted comprehensive preprocessing, addressing challenges like code-switching and use of emoji in tweets. Non-Arabic portions of the tweets were then automatically translated into Arabic. Instead of removing emojis and hashtags, these were converted into Arabic descriptive text to preserve the sentiment of the tweets. For Subtask 2A, the team used AraBERT-Covid19\footnote{\url{https://huggingface.co/moha/arabert_arabic_covid19}} with hyperparameters optimized through the optimization framework Optuna. As for Subtask 2B, a soft voting ensemble method is used with five optimized AraBERTv02-Twitter~\cite{antoun2020arabert} models, each with identical hyperparameters and architecture, only differing by random initialization. AraBERTv02-Twitter was selected since it is based on the effective AraBERT mode, with continued pre-training on $60M$ Arabic tweets, making it suitable for Subtask 2B focused on tweets. 



\paragraph{Frank~\cite{araieval-arabicnlp:2023:bothtasks:frank}} After preprocessing using AraBERT preprocessor, multilingual BERT ~\cite{devlin2018bert} was fine-tuned for Subtask 1A, and MARBERT was fine-tuned for Subtask 2A. 

\paragraph{HTE~\cite{araieval-arabicnlp:2023:task1A:HTE}}
Participating in Subtask 1A, the team fine-tuned the MARBERT model in a multitask setting: a primary binary classification task to identify the presence of persuasive techniques in text generally, and an auxiliary task focused on classifying texts based on their type (tweet or news). It was expected that the auxiliary task would help the primary task in learning specific lexical and syntactic information about tweets or news related to persuasive content. Given the imbalance in the dataset, the team employed focal loss to optimize both tasks. On the test set, the system ranked highest on the leaderboard. 


\paragraph{Itri Amigos~\cite{araieval-arabicnlp:2023:task:Itri_Amigos}}
The team submitted runs for all four subtasks. Preprocessing was applied using AraBERT preprocessor. Further preprocessing was done for all subtasks but 1B, where links and mentions were removed. For subtasks 1A and 1B, the team fine-tuned the AraBERTv2 transformer model. To address the class imbalance in the datasets, class weights were incorporated during training. As for subtasks 2A and 2B that are mainly targeting tweets, AraBERTv02-Twitter was fine-tuned for the tasks.

\paragraph{KnowTellConvince~\cite{araieval-arabicnlp:2023:taskboth:KnowTellConvince}} The team participated in subtasks 1A, 2A and 2B using an ensemble of the following four models. \emph{(i)} fine-tuned BERT Arabic base model~\cite{safaya-etal-2020-kuisail} with a contrastive loss function; \emph{(ii)} fine-tuned BERT Arabic base model with a cross entropy loss function; \emph{(iii)} fine-tuned BERT Arabic base on XNLI dataset to capture nuances relevant to sentiment as part of the system architecture; and \emph{(iv)} a model utilizing sentence embeddings from BERT Arabic base followed by computing cosine similarity between pairs of sentences from the data, that finally goes through Gaussian Error Linear Unit (GELU) activation.

\paragraph{Legend~\cite{araieval-arabicnlp:2023:task:legend}} team participated in Task 1, in which XLM-RoBERTa was implemented. To address the class imbalance in the dataset, the team adjusted the learning process using class weights. A learning rate scheduler was implemented to dynamically adjust the learning rate during training. Specifically, they used a StepLR scheduler with a reduction factor of 0.85 applied every 2 epochs. This scheduling strategy contributes to the training stability and the controlled convergence.

\paragraph{Mavericks~\cite{araieval-arabicnlp:2023:bothtasks:Mavericks}}

Targeting subtasks 1A and 2A, several transformer-based models were fine-tuned on the provided dataset. The models include: AraBERT, MARBERT and AraELECTRA~\cite{antoun-etal-2021-araelectra}. Ensembling was utilized using hard voting, where the majority vote of all the predictions is selected as the final prediction. 

\paragraph{Nexus~\cite{araieval-arabicnlp:2023:bothtasks:nexus}}
The team explored performance of fine-tuning several pre-trained language models (PLMs) including AraBERT, MARBERT, and QARiB in subtasks 1A and 2A. In addition to that, experiments with GPT-4~\cite{openai2023gpt4} in both zero-shot and few-shot settings were conducted for both subtasks. Performance of the GPT-4 model was notably lower than the fine-tuned models.

\paragraph{PD-AR~\cite{araieval-arabicnlp:2023:task:PD-AR}}
For both sub-tasks 2A and 2B, the team employed the AraBERTv0.2-Twitter-base model and utilized the provided training and development sets to train the model. Before training, some preprocessing of the text was performed. Compared to fine-tuning several other PLMs such as XLM-RoBERTa~\cite{conneau2020unsupervised}, the Arabic-specific model showed significantly improved performance. 


\paragraph{PTUK-HULAT~\cite{araieval-arabicnlp:2023:task:PTUK-HULAT}}
The team participated in Subtask 2A, in which they fine-tuned a multilingual DistilBERT model on the corresponding binary classification data. They then used the fine-tuned model to predict whether a tweet is dis-informative or not.

\paragraph{Raphael~\cite{araieval-arabicnlp:2023:task1:raphael}}
For both subtasks 1A and 1B, they used MARBERT as the encoder. In addition to that, they used GPT-3.5~\cite{brown2020language} in order to generate English descriptions of the Arabic texts and to provide tone and emotional analysis. The resulting English text and tone descriptions were then encoded using RoBERTa~\cite{liu2019roberta} and were further concatenated to the MARBERT encodings. Finally, the full embeddings were passed to a binary classification head and to multilabel classification heads for Subtasks 1A and 1B, respectively.

\paragraph{ReDASPersuasion~\cite{araieval-arabicnlp:2023:task:redaspersuasion}} The initial structure of the system has three main components:
\emph{(i)} A multilingual transformer model that tokenizes the input and produced a [CLS] embedding output;
\emph{(ii)} A feature engineering module designed to extract language-agnostic features for persuasion detection;
\emph{(iii)} A multi-label classification head that integrates the first and the second components, using a sigmoid activation and cross entropy loss. For subtasks 1A and 1B, the system was paired with DistilBERT~\cite{sanh2019distilbert} for the official submission, but follow-up experiments for Subtask 1A showed that using XLM-RoBERTa, yielded the best Micro F1 score on test.


\paragraph{rematchka~\cite{araieval-arabicnlp:2023:bothtasks:rematchka}}
For all subtasks, ARBERTv2~\cite{abdul-mageed-etal-2021-arbert}, AraBERTv2, and MARBERT models were trained on the provided datasets. For Subtask 1A, different techniques such as fast gradient methods and contrastive learning were applied. Moreover, the team employed back-translation between Arabic and English for data augmentation. As for Subtask 1B, different loss functions, including Asymmetric loss and Distribution Balanced loss were tested. Moreover, a balanced data-sampler for multilabel datasets was used. Fro both subtasks, prefix tuning was used for model training.

\paragraph{UL \& UM6P~\cite{araieval-arabicnlp:2023:task:UL_UM6P}} 
used an Arabic pre-trained transformer combined with a classifier. The performance of three transformer models was evaluated for sentence encoding. For Subtask 1A, the MARBERTv2 encoder was used, and the model was trained with cross-entropy and regularized Mixup (RegMixup) loss functions. For Subtask 1B, the AraBERT-Twitter-v2 encoder was used, and the model was trained with the asymmetric multi-label loss. The significant impact of the training objective and text encoder on the model's performance was highlighted by the results. For Subtask 2A, the AraBERT-Twitter-v2 encoder was used, and the model was trained with cross-entropy loss. For Subtask 2B, the MARBERTv2 encoder was used, and the model was trained with the Focal Tversky loss.



\paragraph{USTHB~\cite{araieval-arabicnlp:2023:task2:usthb}} For both subtasks 2A and 2B, the system start with extensive preprocessing of the data. Then, the FastText model is used for feature extraction in addition to TF-IDF to vectorize the data. SVM was then trained as a classifier. 

%% file: sections/conclusion.tex
\section{Conclusion and Future Work}
\label{sec:conclusion}
We presented an overview of the \areval{} shared task at the ArabicNLP 2023 conference, targeting two shared tasks: {\em (i)} persuasion technique detection, and {\em (ii)} disinformation detection. The task attracted the attention of many teams: a total of 25 teams registered to participate during the evaluation phase, with 14 and 16 teams eventually making an official submission on the test set for tasks 1 and 2, respectively. Finally, 17 teams submitted a task description paper. Task 1 aimed to identify the propaganda techniques used in multi-genre text snippets, including tweets and news articles, in both binary and multilabel settings. On the other hand, Task 2 aimed to detect disinformation in tweets in both binary and multiclass settings. For both tasks, the majority of the systems fine-tuned pre-trained Arabic language models and used standard pre-processing. Several systems explored different loss functions, while a handful of systems utilized data augmentation and ensemble methods.

Given the success of the task this year, we plan to run a future edition with an increased data size, and with wider coverage of domains, countries, and Arabic dialects. We are also considering implementing a multi-granularity persuasion techniques detection setting.

\section*{Limitations}
Task 1 was limited to binary an multilabel classification. A natural next step would have been to also run a span detection subtask, which is a more complex task. This was left for future editions of \areval. This is to ensure enough participation after building a strong community working on propaganda detection over Arabic content in the less complex setups. As for Task 2, we observe the systems achieved significantly high performance, even in the more challenging multiclass setup. One potential reason might be that the dataset developed was too easy. Investigating how to make this task more challenging while reflecting real-world scenarios was not in this edition of the shared task, but is within our future plan.